\documentclass[10pt,twocolumn,letterpaper]{article}

\usepackage{arxiv}
\usepackage{times}
\usepackage{epsfig}
\usepackage{graphicx}
\usepackage{amsmath}
\usepackage{amssymb}

\usepackage{times}
\usepackage{mathrsfs}
\usepackage{amsmath}
\usepackage{amsfonts}
\usepackage{epstopdf} % for eps
\usepackage{xcolor}

\usepackage[pagebackref=true,breaklinks=true,letterpaper=true,colorlinks,bookmarks=false]{hyperref}

\cvprfinalcopy % *** Uncomment this line for the final submission

\def\cvprPaperID{927} % *** Enter the CVPR Paper ID here

\ifcvprfinal\fi
\usepackage{fancyhdr}
\begin{document}

%%%%%%%%% TITLE
\title{Unsupervised Single Image Deraining with Self-supervised Constraints}

% \title{UD-GAN: Unsupervised Deraining Generative Adversarial Network with Self-supervision}

% 2.Supervised or Unsupervised? Learning to Remove Rain from Image by Self-supervision

% 3.Unsupervised Image Deraining with Self-supervision

% 4.Learning and Using the Self-supervision for Single Image Deraining

% 5.Self-supervised Learning of Raw Data for Single Image Deraining

% 6.Deep Self-supervised Rain Removal in Images

\author{Xin Jin, Zhibo Chen, Jianxin Lin, Zhikai Chen, Wei Zhou\\
University of Science and Technology of China\\
jinxustc@mail.ustc.edu.cn, chenzhibo@ustc.edu.cn, \{linjx, czk654, weichou\}@mail.ustc.edu.cn
}

\maketitle
%\thispagestyle{empty}

%%%%%%%%% ABSTRACT
\begin{abstract}

    Most existing single image deraining methods require learning supervised models from a large set of paired synthetic training data, which limits their generality, scalability and practicality in real-world multimedia applications. Besides, due to lack of labeled-supervised constraints, directly applying existing unsupervised frameworks to the image deraining task will suffer from low-quality recovery. Therefore, we propose an Unsupervised Deraining Generative Adversarial Network (UD-GAN) to tackle above problems by introducing self-supervised constraints from the intrinsic statistics of unpaired rainy and clean images. Specifically, we firstly design two collaboratively optimized modules, namely Rain Guidance Module (RGM) and Background Guidance Module (BGM), to take full advantage of rainy image characteristics: The RGM is designed to discriminate real rainy images from fake rainy images which are created based on outputs of the generator with BGM. Simultaneously, the BGM exploits a hierarchical Gaussian-Blur gradient error to ensure background consistency between rainy input and de-rained output. Secondly, a novel luminance-adjusting adversarial loss is integrated into the clean image discriminator considering the built-in luminance difference between real clean images and de-rained images. Comprehensive experiment results on various benchmarking datasets and different training settings show that UD-GAN outperforms existing image deraining methods in both quantitative and qualitative comparisons.

\end{abstract}

\section{Introduction}\label{sec:introduction}

\begin{figure}
  \centerline{\includegraphics[width=1.0\linewidth]{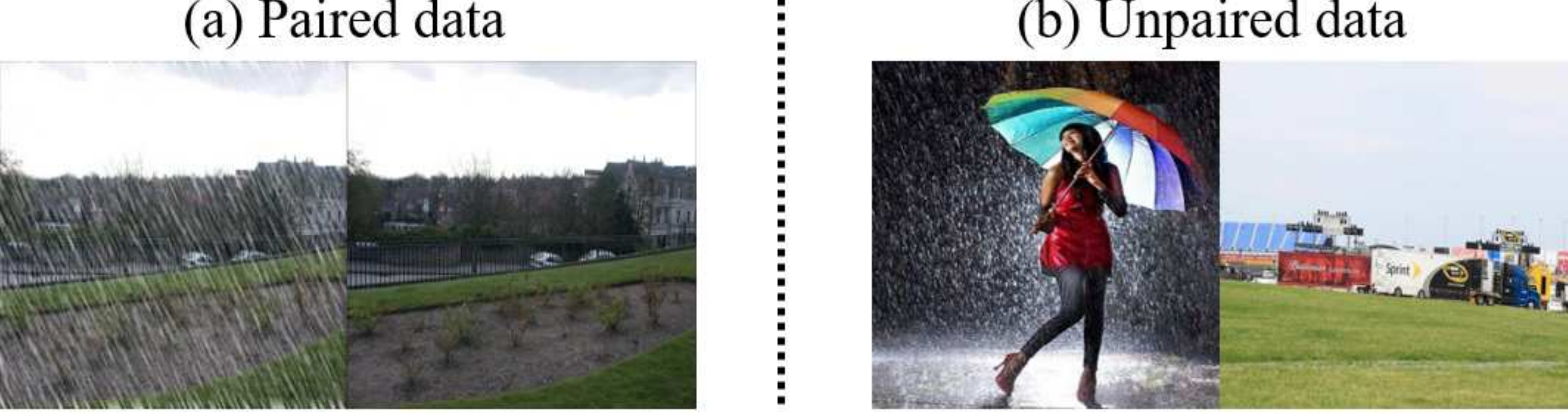}}
  \caption{(a) Synthetic rainy image and corresponding clean image. (b) Real-world rainy image and random clean image.}
  \centering
\label{fig:paired-unpaired}
\end{figure}

Single image deraining is important for many outdoor multimedia applications such as surveillance, pedestrian detection and autonomous driving, etc. Recently, many deep learning-based deraining methods have been proposed to address this problem \cite{he2016deep,fu2017clearing,fu2017removing,yang2017deep,zhang2018density,li2018recurrent,fan2018residual,li2018non}. These methods are mainly trained on synthetic rainy-clean image pairs (Figure \ref{fig:paired-unpaired} (a)) in a supervised manner and then applied in real-world rainy scenarios, which causes several limitations: (1) The manually synthetic rain shapes usually differ from the real ones in nature due to the rain distribution gap between them, which causes these fully-supervised deraining approaches to have limited ability to remove unknown rain from real-world rainy images. (2) The pattern and style of rain are various, which makes it difficult to treat all scenarios with just one fully-supervised deraining model. For example, the model trained on the heavy rain could not be directly applied in the light-rainy scenario, and vice verse. Therefore, we attempt to handle with the deraining task from a completely different perspective of resorting to unsupervised learning with unpaired real-world data (Figure \ref{fig:paired-unpaired} (b)), so that the problems mentioned above could be well solved.

\begin{figure}
  \centerline{\includegraphics[width=1.0\linewidth]{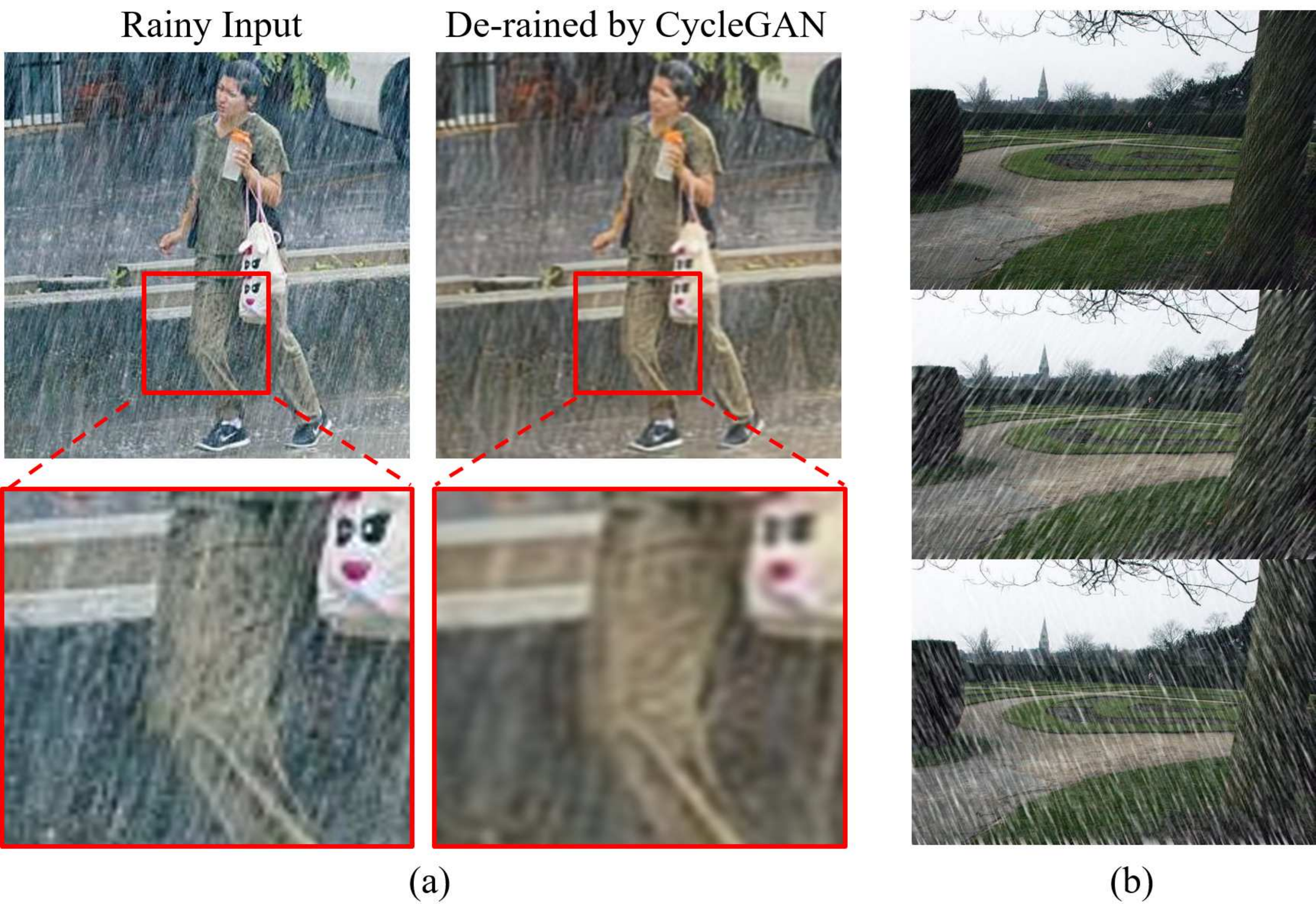}}
  \caption{(a) Directly using CycleGAN \cite{zhu2017unpaired} to do image deraining will leave lots of residual rain streaks and generate unacceptable artifacts. (b) The direction, shape and density of rain streaks are various.}
  \centering
\label{fig:Shown}
\end{figure}

Unfortunately, although significant success has been achieved in unsupervised learning-based image processing models, such as CycleGAN \cite{zhu2017unpaired} and WESPE \cite{ignatov2017wespe}, they still fail to surpass previous supervised models in deraining task. There exist two major reasons: (1) Unsupervised training schemes usually suffer from the under-constrained problem since supervised constraints such as mean-square error (MSE) between the output and the ground truth cannot be applied directly, which often results in unwanted artifacts as shown in Figure \ref{fig:Shown} (a). (2) Most unsupervised networks are designed to learn a one-to-one transformation, such as horse-to-zebra, day-to-night, etc. But for deraining, they hardly capture the varied transformations from rainy to clean because the direction, shape and density of rain streaks are various as shown in Figure \ref{fig:Shown} (b).

To address these two problems, in this paper, we propose a novel perspective to achieve unsupervised deraining: \emph{Instead of relying solely on pure unsupervised domain transformations, we introduce self-supervised constraints from the intrinsic statistics of unpaired rainy and clean images to guide deraining, which solves the first problem of under-constraint.} Specifically, we propose an end-to-end learning Unsupervised Deraining Generative Adversarial Network (UD-GAN) with two collaboratively optimized modules: Rain Guidance Module (RGM) and Background Guidance Module (BGM). RGM indirectly constrains the solution space of generated de-rained images by constraining the difference (i.e. removed rain streaks) between rainy input and de-rained output, \emph{which further solves the second negative influence on unsupervised deraining due to the variety of rain streaks}. BGM ensures background consistency by imposing a hierarchical Gaussian-Blur gradient error between rainy input and de-rained output. Considering the built-in luminance difference between real clean images and de-rained images, a luminance-adjusting adversarial loss is designed for obtaining more natural and realistic de-rained results. The contributions of this paper are summarized as follows:

\begin{itemize}
    \item To the best of our knowledge, this study is the first data-driven attempt to unsupervised learning for deraining task trained with unpaired image sets.

    \item By learning from the intrinsic statistics of raw data in a self-supervised manner, we provide a novel perspective for unsupervised training, which opens a new way to single image deraining, bringing it closer to practical applications.

    \item Extensive performance evaluation on both synthetic and real-world datasets validates the effectiveness of our method. Especially for the improvement of subjective effects in real-world rainy scenes, UD-GAN greatly exceeds existing supervised methods and can be easily generalized to other computer vision tasks.
\end{itemize}

\begin{figure*}
  \centerline{\includegraphics[width=0.87\linewidth]{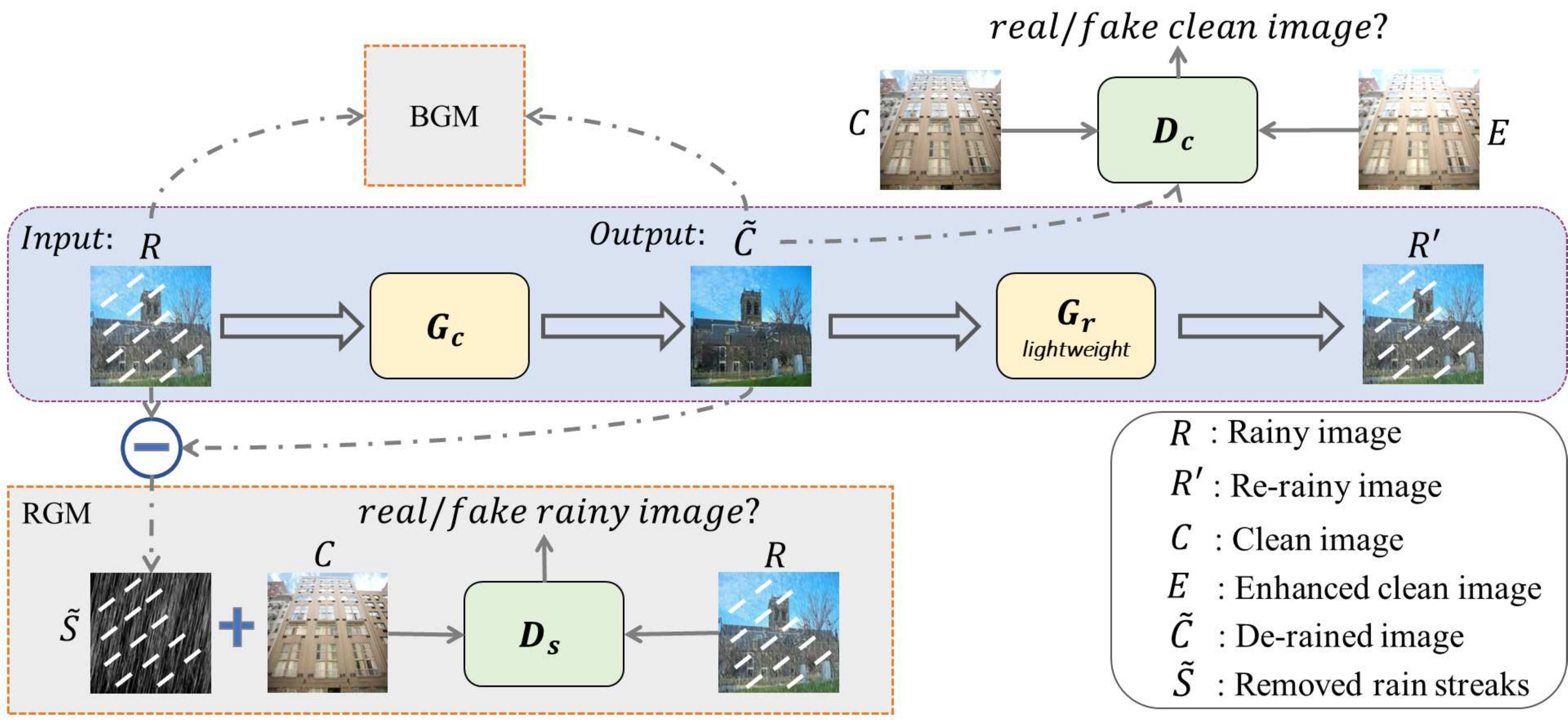}}
  \caption{An overview of the proposed UD-GAN with two collaboratively optimized Rain Guidance Module (RGM) and Background Guidance Module (BGM). The RGM indirectly helps the generator $G_c$ de-rain the rainy input by using discriminator $D_s$ to constrain the removed rain streaks, while the BGM ensures the background consistency between rainy input and de-rained output. $D_c$ is a clean image discriminator with luminance adjustment function. The lightweight rain-added generator $G_r$ is to avoid style/color variation.}
  \centering
\label{fig:UD-GAN}
\end{figure*}

\section{Related Work}\label{sec:related-work}

\subsection{Single Image Deraining}

% Unlike video-based methods \cite{ren2017video,li2018video,chen2018robust,liu2018erase,wei2017should} that leverage temporal information in rain removal,

\textbf{Traditional Methods:} Traditional prior-based methods have been proposed in the literature to deal with single image deraining problem, such as sparse coding-based methods \cite{huang2014self,luo2015removing,zhu2017joint}, low-rank representation-based methods \cite{chen2013generalized,zhang2017convolutional} and gaussian mixture model-based (GMM) methods \cite{li2016rain,li2017single}, etc. The main limitation of existing prior-based methods is that they often tend to have under de-rained effect by leaving residual rain streaks \cite{kang2012automatic} or over-smooth image details \cite{luo2015removing}.

\textbf{Deep Neural Network (DNN):} The renaissance of DNN remarkably accelerated the progress of deraining task: Fu et al. \cite{fu2017clearing} proposed a learning-based rain removal solution, then they also combined ResNet \cite{he2016deep} and focused on high-frequency details while deraining in DetailsNet \cite{fu2017removing}. \cite{yang2017deep} proposed a deep recurrent network named JORDER to remove rain streaks progressively. This year, \cite{zhang2018density} presented a density-aware multi-stream connected network called DID-MDN for joint rain density estimation and deraining. \cite{li2018recurrent} proposed a recurrent neural network with dilated convolution \cite{yu2015multi} and Squeeze-and-Excitation (SE) blocks \cite{hu2017squeeze}, called RESCAN. \cite{fan2018residual} designed a residual-guide network (RGN) to achieve a coarse-to-fine deraining. \cite{li2018non} introduced a non-locally enhanced encoder-decoder network (NLEDN) for more accurate rain removal. In general, the common idea of above methods is limited to view deraining as a regression problem and learn a mapping between synthetic rainy inputs and ground truths using a CNN structure in a fully-supervised manner, which limits their generality, scalability and practicality in real-world rainy scenes.

\subsection{Unsupervised Learning}

Recently, unsupervised learning-based image processing applications emerged with promising performance \cite{zhang2018deep,wang2018transferable,yuan2018unsupervised,madam2018unsupervised,ren2017unsupervised,guizilini2017unsupervised,yang2017unsupervised,xu2017unsupervised}. These methods usually can be divided into two categories: One is to utilize unsupervised learning to estimate the data distribution for data enhancement, and then use the enhanced data to train the model \cite{chen2018image}, which has limited scalability because the distribution of some data is difficult to be estimated such as rain. Another is to use unpaired data to achieve domain transfer, DualGAN \cite{yi2017dualgan} and CycleGAN \cite{zhu2017unpaired} are two classic GAN-based \cite{goodfellow2014generative} works belonging to this category, and both of them use a pair of GANs to learn the transformation. However, training GANs is highly unstable \cite{salimans2016improved} and thus using two GANs simultaneously escalates in instability. Moreover, it is impossible to directly apply DualGAN or CycleGAN for image deraining, because of the above-mentioned defects existing in unsupervised learning: (1) under-constrained problem and (2) hard to capture the varied transformations from rainy inputs to clean outputs due to rain streaks' diversity.

\subsection{Self-supervised Learning}

The debilitating limitation of supervised learning and the defect of unsupervised learning together necessitate the need for self-supervised learning, which is a form of unsupervised learning where the data provides the supervision. Self-supervised learning has demonstrated success in many computer vision applications \cite{agrawal2015learning,doersch2015unsupervised,owens2016ambient,wang2015unsupervised,stewart2017label,gomez2017self,gidaris2018unsupervised}. Among them, \cite{stewart2017label} introduced a new approach to supervising neural networks without any labeled examples by specifying constraints that are derived from prior domain knowledge, e.g., from known laws of physics. \cite{gomez2017self} put forward an idea of performing self-supervised learning of visual features by mining a large scale corpus of multi-modal (i.e. text and image) documents. \cite{gidaris2018unsupervised} leveraged semantic feature in self-supervised manner to achieve the recognition of 2D image rotation.

Inspired by these works, in this paper, we try to fully exploit the intrinsic characteristics of original data in a self-supervised manner, which achieves the goal of transforming rainy inputs to clean outputs without any paired data.

\section{Unsupervised Deraining Generative Adversarial Network (UD-GAN)}\label{sec:UD-GAN}

% \subsection{UD-GAN Framework}

As illustrated in Figure \ref{fig:UD-GAN}, our goal is learning to transform images from the rainy domains $R$ to the target clean domain $C$ given random and unpaired training samples ${\{ {r_i}\} _{i = 1}^N \in R}$ and ${\{ {c_j}\} _{j = 1}^M \in C}$. The first generator $G_c$ is used to transform rainy inputs to the clean outputs, which captures a deraining transformation: $R$ $\rightarrow$ $C$. Correspondingly, an adversarial discriminator $D_c$ is used to distinguish between real clean images ${C}$ and fake de-rained images ${\tilde{C}}$, where ${\{ {\tilde{c}}\} \in {\tilde{C}}}$ represents a fake de-rained result:

\begin{equation}
\tilde{c} = G_c(r), \hspace{0.3cm} D_c(c, \tilde{c}) ?= real/fake.
\label{eq:1}
\end{equation}

However, training deraining generator $G_c$ with adversarial cost alone may introduce visual artifacts \cite{isola2017image} in certain regions of the generated de-rained output, but the clean image discriminator $D_c$ can still end up classifying it as real data rather than generated data, which is unacceptable. To solve this problem, self-supervised constraints from the intrinsic statistics of deraining problem are introduced in the following sections.

% Hence, to solve this problem and further promise the whole image consistency \cite{zhu2017unpaired}, we employ the second lightweight generator $G_r$ to learn the inverse rain-added transformation:

% \begin{equation}
% r' = G_r(\tilde{c}),
% \label{eq:2}
% \end{equation}
% where ${\{ {r'}\} \in {R'}}$, representing the reconstructed re-rainy results. It is worth noting that the rain-added generator $G_r$ is a lightweight network compared to deraining generator $G_c$ and there is also no corresponding discriminator, that's because training a pair of GANs simultaneously is highly unstable \cite{salimans2016improved}, so we just use a single generator $G_r$ to promise the whole image consistency and optimize it only through the following introduced Consistency Loss $\mathcal{L}_{cyc}$.

\subsection{Self-supervision by Rainy Image}

\subsubsection{Rain Guidance Module (RGM)}

As discussed above, existing unsupervised frameworks tend to suffer from the under-constrained problem due to lack of labeled-supervised constraints and hard to capture the varied transformations from rainy inputs to clean outputs due to the diversity of rain streaks. Hence, we introduce a Rain Guidance Module (RGM) to take full advantage of the intrinsic statistics of original rainy inputs, and in turn use the learned rain characteristics to guide in better deraining.

In detail, based on the widely used rain model \cite{huang2012context,luo2015removing,li2016rain,zhu2017joint,jindecomposed,zhang2018density,wang2018rain}: ${R=C+S}$, we leverage the inner interdependency between clean images $C$ and rain streaks $S$ to indirectly constrain the de-rained outputs $\tilde{C}$ of the deraining generator $G_c$. Refer to Figure \ref{fig:UD-GAN}, RGM mainly depends on an independent discriminator $D_s$. We first obtain the difference (i.e. removed rain streaks $\tilde{S}$) between rainy inputs $R$ and de-rained outputs $\tilde{C}$ by ${\tilde{S}=R-\tilde{C}}$. Then we superimpose rain streaks $\tilde{S}$ on the real clean images $C$ to create fake rainy images $\tilde{S}+C$, and distinguish them from real rainy images $R$ by rain streaks discriminator $D_s$:

\begin{equation}
\begin{aligned}
\tilde{s} = r - \tilde{c}, \hspace{0.5cm} D_s(r, \tilde{s}+c) ?= real/fake,
\label{eq:3}
\end{aligned}
\end{equation}
where ${\{ {\tilde{s}}\} \in \tilde{S}}$, ${\{ {\tilde{c}}\} \in \tilde{C}}$, ${\{ {r}\} \in R}$ and ${\{ {c}\} \in C}$. We hold the view that if the deraining generator $G_c$ is trained well enough, when we superimpose the removed rain streaks $\tilde{S}$ on any real clean image $C$, the obtained fake rainy images $\tilde{S}+C$ should be indistinguishable from the real rainy images $R$. That is, RGM authentically acts as a "supervising teacher" to indirectly help the deraining generator $G_c$ obtain better de-rained outputs $\tilde{C}$ by encouraging the removed rain streaks $\tilde{S}=R-\tilde{C}$ to be close to the real rain streaks.

Correspondingly, a rain guidance loss $\mathcal{L}_{guid}^{R}$ is defined as follows, which optimizes the deraining generator $G_c$ and rain streaks discriminator $D_s$ simultaneously:
\begin{equation}
\begin{aligned}
{\mathcal{L}_{guid}^{R}}({G_c},{D_s}) = {\mathbb{E}_{r\sim{p_{data}}(r)}}[\log{D_s}(r)] \\
+ {\mathbb{E}_{r,c\sim{p_{data}}(r,c)}}[\log(1 - {D_s}((r-G_c(r)+c)))].
\end{aligned}
\label{eq:6}
\end{equation}

This guidance from RGM essentially can be viewed as a kind of adversarial collaboration: the more realistic the removed rain streaks $\tilde{S}$ are, the cleaner the de-rained results $\tilde{C}$ are, and vice versa.

\subsubsection{Background Guidance Module (BGM)}

In addition to the correct transformation from rainy to clean domain, another important goal in image deraining is to ensure content consistency and avoid losing important details after deraining. To achieve this goal, we design another Background Guidance Module (BGM) to further utilize the input rainy images to provide more reasonable and reliable self-supervised constraints for the deraining generator $G_c$.

Inspired by \cite{kang2012automatic,wang2017hierarchical,fu2017clearing,fu2017removing}, we found that after applying an appropriate low-pass filter such as Gaussian blur kernel \footnote{http://homepages.inf.ed.ac.uk/rbf/HIPR2/log.htm} or Guided Filtering \cite{he2013guided}, low-pass versions of both the rainy image and the clean image are rain-free and approximately equal, they only contain the background/content features. We can take advantage of this fact to guide the deraining process. Specifically, we hierarchically use the Gaussian blur kernels with different scales $\sigma$ to filter rainy input $R$ and de-rained output $\tilde{C}$, obtaining their background features respectively. Figure \ref{fig:BGM} shows that as we increase in Gaussian blur scale $\sigma$, not only the background features start to look alike, but also the average gradient error between them decreases. Based on this, we naturally form a guidance for the deraining generator $G_c$ through a background guidance loss $\mathcal{L}_{guid}^{B}$, which enforces the Gaussian-Blur gradients of rainy input and de-rained output to match at different scales $\sigma$:

\begin{figure}
  \centerline{\includegraphics[width=0.8\linewidth]{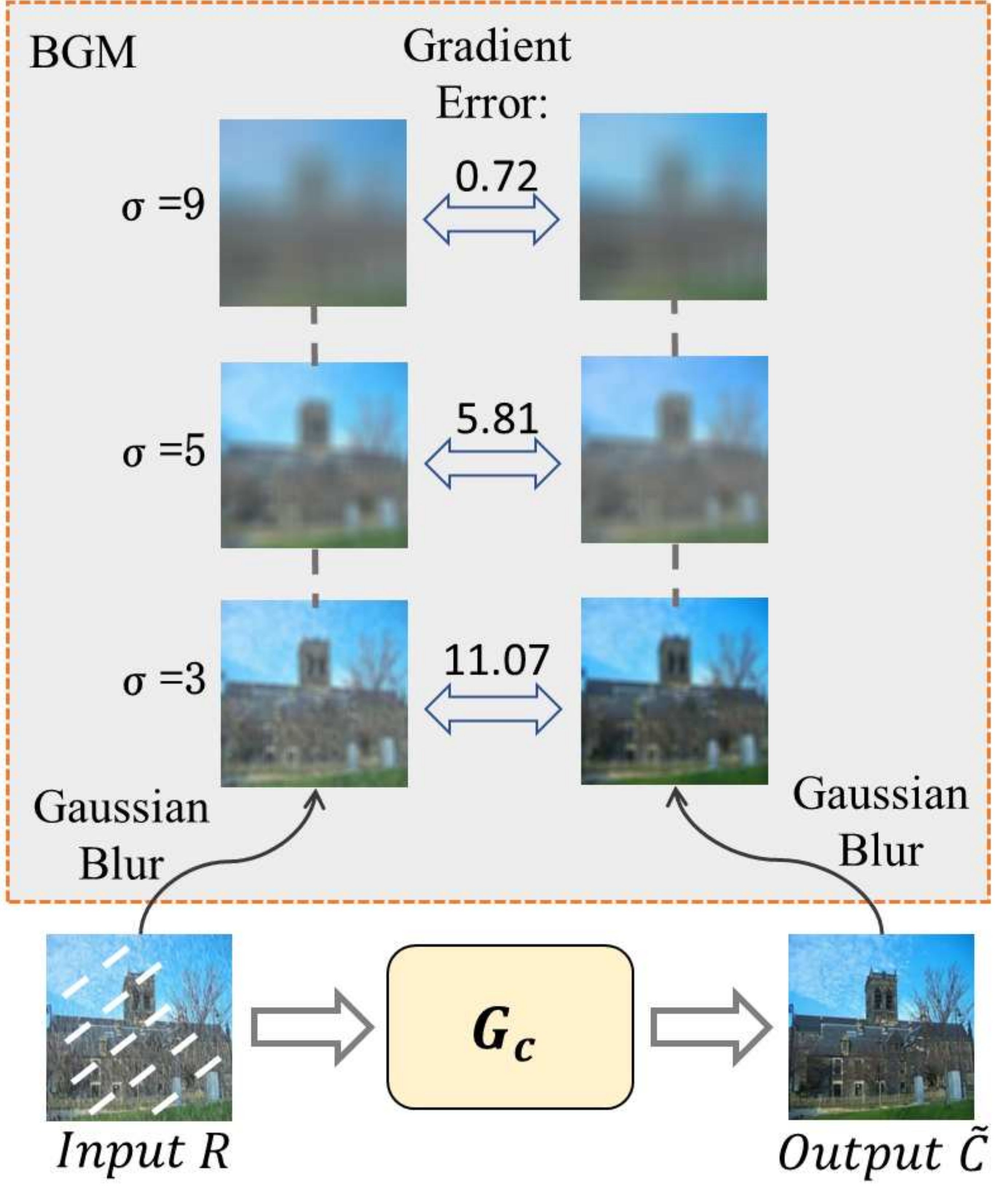}}
  \caption{Background Guidance Module (BGM), which hierarchically uses the gradient errors at different Gaussian-Blur levels to ensure the content consistency between rainy inputs $R$ and de-rained outputs $\tilde{C}$.}
  \centering
\label{fig:BGM}
\end{figure}

\begin{equation}
{\mathcal{L}_{guid}^{B}(G_c)} = {\sum_{\sigma=3,5,9}\lambda_{\sigma}}|\nabla{B_{\sigma}}(r)- \nabla{B_{\sigma}}(G_c(r))|,
\label{eq:8}
\end{equation}
where $\nabla$, $B_{\sigma}$ denote the gradient computation and Gaussian blur operation. $\lambda_{\sigma}$ values are used to balance errors at different Gaussian-Blur levels. We believe that it is necessary to utilize these gradient errors hierarchically, because different levels of background features contain different levels of important detail information. Hence, based on experimental attempts, we set $\lambda_{\sigma}$ as [0.01, 0.1, 1] for $\sigma = {3,5,9}$, respectively.

\subsection{Self-supervision by Clean Image}
% Loss Function}

\begin{figure}
  \centerline{\includegraphics[width=1.0\linewidth]{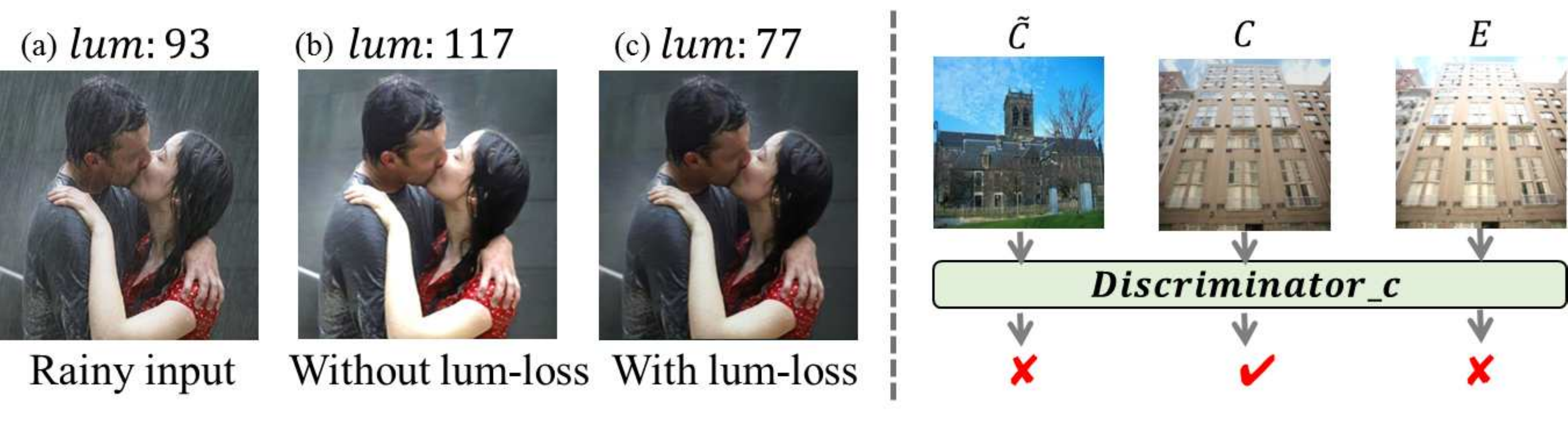}}
  \caption{Left: directly using the real clean images (usually sunny in dataset) to constraint the de-rained outputs often causes over-bright results (b) and don't match the cloudy situation (c). Right: by enhancing luminance in clean images ${\{ {c_j}\} _{j = 1}^M \in C}$, we additionally generate a negative training sample set $E$: ${\{ {e_k}\} _{k = 1}^K \in E}$.}
  \centering
\label{fig:LPS}
\end{figure}

As shown in Eq. \ref{eq:1}, the deraining generator $G_c$ and the clean image discriminator $D_c$ form a complete GAN \cite{goodfellow2014generative}, an adversarial loss should be applied to them, its value indicates what extent the de-rained output of deraining generator $G_c$ looks like a clean image. However, we observe that simply training $D_c$ through a standard adversarial loss like in \cite{goodfellow2014generative,zhu2017unpaired,isola2017image} to separate generated de-rained images and true clean images is not sufficient due to the built-in luminance difference between them. In detail, we find that (1) the real rainy images are almost cloudy, their average luminance is usually lower. If we directly use the real clean images (usually sunny in our collected real-world dataset) to constraint the de-rained outputs through a standard adversarial loss, the de-rained outputs are often too bright and do not match the cloudy day (Figure \ref{fig:LPS} (left)), (2) the rain streaks of the synthetic rainy image always appear as some bright lines (Figure \ref{fig:paired-unpaired} (a)). So, those de-rained images with higher illuminace may leave more residual rain streaks.

%so a de-rained output image with higher illuminance will make the residual rain streaks more obvious, which is unexpected.
% , a visual picture depicting this effect is shown in Figure \ref{fig:LPS} (left).

We explore to leverage some "clean image constraints" to circumvent the above problems. Specifically, from the clean training samples ${\{ {c_j}\} _{j = 1}^M \in C}$, we additionally generate a negative training sample set $E$: ${\{ {e_k}\} _{k = 1}^K \in E}$ by enhancing the luminance \cite{guo2017lime} of the images in $C$. As shown in Figure \ref{fig:LPS} (right), during training, the clean image discriminator $D_c$ should maximize the probability of assigning the correct label to fake de-rained images $\tilde{C}$, real clean images $C$ and luminance-enhanced clean images $E$, such that the deraining generator $G_c$ can be guided correctly in transforming the rainy input to the clean output. Therefore, we re-define a luminance-adjusting adversarial loss as follows:

\begin{equation}
\begin{aligned}
{\mathcal{L}_{lum\_adv}}({G_c},{D_c}) &= {\mathbb{E}_{c\sim{p_{data}}(c)}}[\log{D_c}(c)] \\
&+ {\mathbb{E}_{e\sim{p_{data}}(e)}}[\log(1 - {D_c}(e))] \\
&+ {\mathbb{E}_{r\sim{p_{data}}(r)}}[\log(1 - {D_c}({G_c}(r)))]. \\
\end{aligned}
\label{eq:5}
\end{equation}

We discuss that adding such a luminance-adjusting adversarial loss ensures that (1) the de-rained results in real-world rainy scene have more realistic and natural luminance, (2) the de-rained results for synthetic rainy images leave less residual rain streaks. Figure \ref{fig:Syn-Real-results},\ref{fig:AblationStudy} in the experiment section validate the improvement of subjective effects.

\subsection{Loss Function}

Except for the above-mentioned rain guidance loss $\mathcal{L}_{guid}^{R}$, background guidance loss $\mathcal{L}_{guid}^{B}$ and luminance-adjusting adversarial loss $\mathcal{L}_{lum\_adv}$, the final loss function for UD-GAN also contains a cycle consistency loss $\mathcal{L}_{cyc}$ to cope with substantial style/color variations between rainy input and de-rained output.

Inspired by \cite{zhu2017unpaired,ignatov2017wespe}, we employ another generator $G_r$ to learn the inverse rain-added transformation: $r' = G_r(\tilde{c})$, where ${\{ {r'}\} \in {R'}}$ representing the reconstructed re-rainy results as shown in Figure \ref{fig:UD-GAN}. Considering training complexity and time complexity, the rain-added generator $G_r$ is designed as a lightweight network compared to the deraining generator $G_c$ and doesn't have corresponding discriminator. We optimize $G_r$ only through a consistency loss:

\begin{equation}
\begin{aligned}
{\mathcal{L}_{cyc}}(G_c,G_r) &= {\mathbb{E}_{r\sim{p_{data}}(r)}}[\|G_r(G_c(r))-r\|_1].
\end{aligned}
\label{eq:6}
\end{equation}

In summary, the final loss for UD-GAN is a weighted sum of the above four losses, where $w_1 \sim w_4$ are the weights to balance different losses. Based on experimental attempts, we finally set $w_1 \sim w_4$ as 1, 5, 1, 0.5:

\begin{equation}
\begin{aligned}
{\mathcal{L}_{total}} &= w_1{\mathcal{L}_{guid}^{R}}
+ w_2{\mathcal{L}_{guid}^{B}}
+ w_3{\mathcal{L}_{lum\_adv}}
+ w_4{\mathcal{L}_{cyc}}.
\end{aligned}
\label{eq:10}
\end{equation}

\subsection{Network Architecture and Training Details}

% For the generators $G_c$ and $G_r$, we use 3, 1 dilated convolution layers \cite{yu2015multi} at the head and tail, and 6, 2 residual blocks \cite{he2016deep} in the middle respectively. Leaky ReLU layer and a batch normalization layer \cite{ioffe2015batch} are added. For the discriminators, we use a 70*70 Markovian Patch-GAN architecture as explored in \cite{li2016precomputed} for $D_c$ and $D_s$.

Source codes are coming soon. During training, the weights of our model are all initialized using the technique described in \cite{he2015delving}. We adopt an Adam \cite{kingma2014adam} solver to minimize the whole objective function in Eq. \ref{eq:10}. We set the mini-batch size to 1, the size of the input image is 512*512. The entire network is trained on two Nvidia 1080Ti GPUs based on Pytorch framework. The initial learning rate $\alpha$ is 0.0002 and decreases as the number of iterations increases. We train the model over 200,000 iterations, until it well converges.

\begin{table*}
\centering
\footnotesize
\caption{Quantitative comparison with existing methods on Rain800, Rain12, Rain100L and Rain100H. The three best performing methods are marked in \textcolor{red}{red}, \textcolor{blue}{blue}, and \textcolor{green}{green}, respectively.}
\label{Quantitative Evaluation}
\begin{tabular}{|c|c|c|c|c|c|c|c|c|c|c|c|c|c|}
\hline
Dataset         & \multicolumn{2}{c|}{Rain800}                                                & \multicolumn{2}{c|}{Rain12}                                                & \multicolumn{2}{c|}{Rain100L}                                              & \multicolumn{2}{c|}{Rain100H}                                              \\ \hline
Metric          & PSNR                                 & SSIM                                 & PSNR                                & SSIM                                 & PSNR                                & SSIM                                 & PSNR                                & SSIM                                 \\ \hline
ID \cite{kang2012automatic}             & 18.88                                & 0.5832                               & 27.21                               & 0.7548                               & 23.13                               & 0.7023                               & 14.02                               & 0.5219                               \\ \hline
DSC \cite{luo2015removing}            & 18.56                                & 0.5996                               & 30.02                               & 0.8745                               & 24.16                               & 0.8721                               & 15.66                               & 0.5467                               \\ \hline
GMM \cite{li2016rain}             & 20.46                                & 0.7297                               & 32.02                               & 0.9155                               & 29.11                               & 0.8809                               & 14.26                               & 0.4225                               \\ \hline
CNN \cite{fu2017clearing}            & 19.22                                & 0.6418                               & 31.19                               & 0.8917                               & 28.70                               & 0.8914                               & 16.08                               & 0.6117                               \\ \hline
DetailsNet \cite{fu2017removing}     & 21.16                                & 0.7320                               & 33.75                               & 0.9319                               & 34.85                               & 0.9508                               & 22.82                               & 0.7409                               \\ \hline
JORDER \cite{yang2017deep}         & 22.29                                & 0.7922                               & 36.02                               & 0.9617                               & 36.11                               & 0.9707                               & 23.45                               & 0.7507                               \\ \hline
DID-MDN \cite{zhang2018density}        & \textcolor{blue}{25.12}                                & \textcolor{blue}{0.8845}                               & 36.14                               & \textcolor{blue}{0.9634}                               & 36.14                               & 0.9711                               & 26.69                               & 0.8774                               \\ \hline
RESCAN \cite{li2018recurrent}         & 24.09                                & 0.8410                               & 35.87                               & 0.9522                               & 36.12                               & 0.9639                               & 26.43                               & 0.8458                               \\ \hline
RGN \cite{fan2018residual}           &  24.04                                & 0.8812                               & 29.45                               & 0.9380                               & 33.16                               & 0.9631                               & 25.25                               & 0.8418                               \\ \hline
NLEDN \cite{li2018non}               & 24.09                                & 0.8766                               & 33.16                               & 0.9192                               & \textcolor{blue}{36.57}                               & \textcolor{blue}{0.9747}                               & \textcolor{blue}{27.03}                       & 0.8819                               \\ \hline
\textbf{UD-GAN$_{syn}$} & \multicolumn{1}{l|}{\textcolor{green}{25.02}} & \multicolumn{1}{l|}{\textcolor{black}{0.8797}} & \multicolumn{1}{l|}{\textcolor{blue}{36.35}} & \multicolumn{1}{l|}{\textcolor{green}{0.9537}} & \multicolumn{1}{l|}{\textcolor{green}{36.20}} & \multicolumn{1}{l|}{\textcolor{green}{0.9723}} & \multicolumn{1}{l|}{\textcolor{green}{26.81}} & \multicolumn{1}{l|}{\textcolor{green}{0.8838}} \\ \hline

\textbf{UD-GAN$_{real}$} & \multicolumn{1}{l|}{\textcolor{black}{24.78}} & \multicolumn{1}{l|}{\textcolor{green}{0.8817}} & \multicolumn{1}{l|}{\textcolor{green}{36.21}} & \multicolumn{1}{l|}{\textcolor{black}{0.9481}} & \multicolumn{1}{l|}{\textcolor{black}{35.95}} & \multicolumn{1}{l|}{\textcolor{black}{0.9683}} & \multicolumn{1}{l|}{\textcolor{black}{26.61}} & \multicolumn{1}{l|}{\textcolor{blue}{0.8860}} \\ \hline

\textbf{UD-GAN} & \multicolumn{1}{l|}{\textcolor{red}{25.98}} & \multicolumn{1}{l|}{\textcolor{red}{0.9093}} & \multicolumn{1}{l|}{\textcolor{red}{37.13}} & \multicolumn{1}{l|}{\textcolor{red}{0.9644}} & \multicolumn{1}{l|}{\textcolor{red}{37.28}} & \multicolumn{1}{l|}{\textcolor{red}{0.9753}} & \multicolumn{1}{l|}{\textcolor{red}{27.75}} & \multicolumn{1}{l|}{\textcolor{red}{0.8931}} \\ \hline
\end{tabular}
\end{table*}

\section{Experiment}\label{sec:experiment}

\subsection{Dataset and Evaluation Metrics}

We carry out deraining experiments below on four widely used synthetic datasets and a real-world dataset. Rain800 \cite{zhang2018density,li2018recurrent}, Rain12, Rain100L and Rain100H \cite{fu2017removing,yang2017deep} are the synthetic datasets with various synthetic rain streaks. For real-world dataset, we collect 784 real-world rainy images (including some snowing images) from the Internet and the previous studies \cite{fu2017removing,yang2017deep,li2018recurrent}, which are diverse in content and rain. All datasets are divided into training and testing set with a ratio of 7:1. To highlight the generalization ability of our model in both synthetic and real-world rainy scenarios, we introduce three training schemes in total: (1) Training only on synthetic datasets, UD-GAN$_{syn}$, (2) Training only on real-world datasets, UD-GAN$_{real}$, (3) Training on all datasets, UD-GAN. All the schemes are evaluated on both synthetic and real-world test datasets.

Deraining performance on the synthetic data is evaluated in terms of PSNR \cite{huynh2008scope} and SSIM \cite{wang2004image}. Performance on real-world images is evaluated visually since the ground truth images are not available. We compare UD-GAN with the following state-of-the-art methods in the same test environment: image decomposition (ID) \cite{kang2012automatic} (TIP'12), discriminative sparse coding (DSC) \cite{luo2015removing} (ICCV'15), gaussian mixture model (GMM) \cite{li2016rain} (CVPR'16), CNN method (CNN) \cite{fu2017clearing} (TIP'17), DetailsNet \cite{fu2017removing} (CVPR'17), JORDER \cite{yang2017deep} (CVPR'17), DID-MDN \cite{zhang2018density} (CVPR'18), RESCAN \cite{li2018recurrent} (ECCV'18), RGN \cite{fan2018residual} (ACMMM'18) and NLEDN \cite{li2018non} (ACMMM'18).

\begin{figure}
  \centerline{\includegraphics[width=1.0\linewidth]{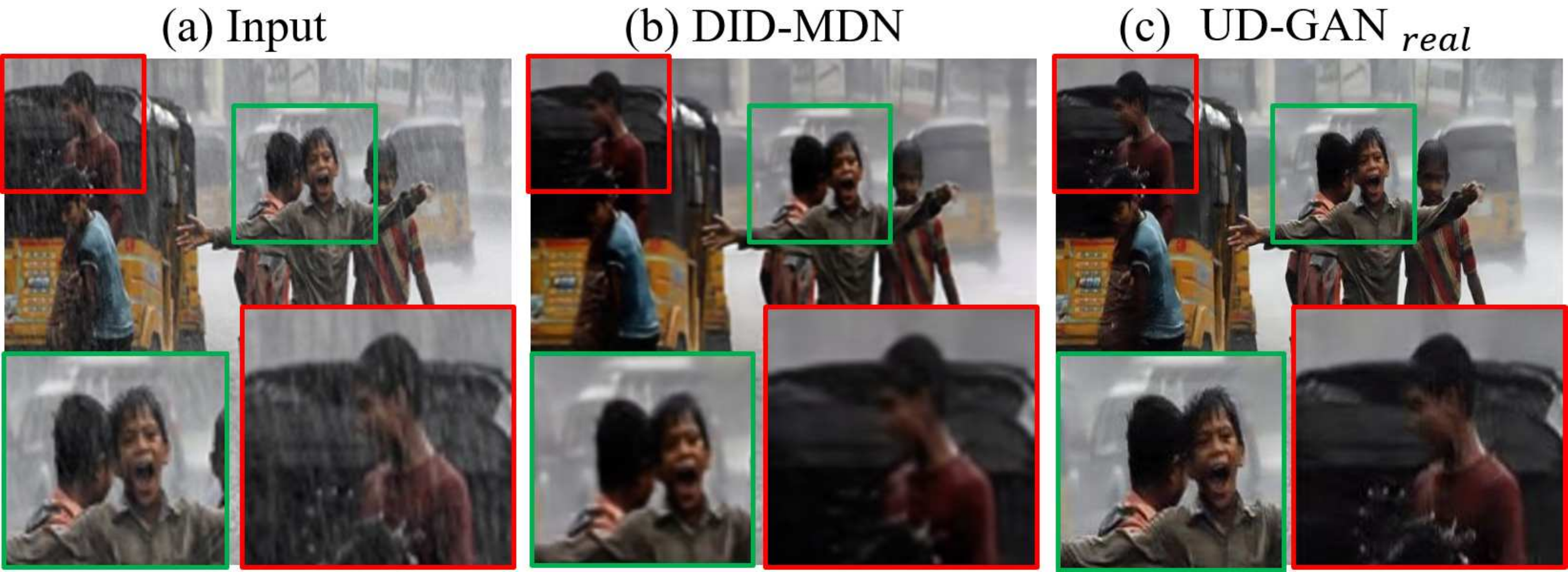}}
  \caption{UD-GAN$_{real}$ that trained only on real rainy data shows a little bit lower PSNR (around 0.4dB) compared with other supervised methods on synthetic datasets, but achieves much better subjective results for real-world rainy images. As shown above, the details of the boy's face and the car's logo have been clearly preserved in result (c), compared with DID-MDN's blur result (b).}
  \centering
\label{fig:real}
\end{figure}

\begin{figure*}
  \centerline{\includegraphics[width=1.0\linewidth]{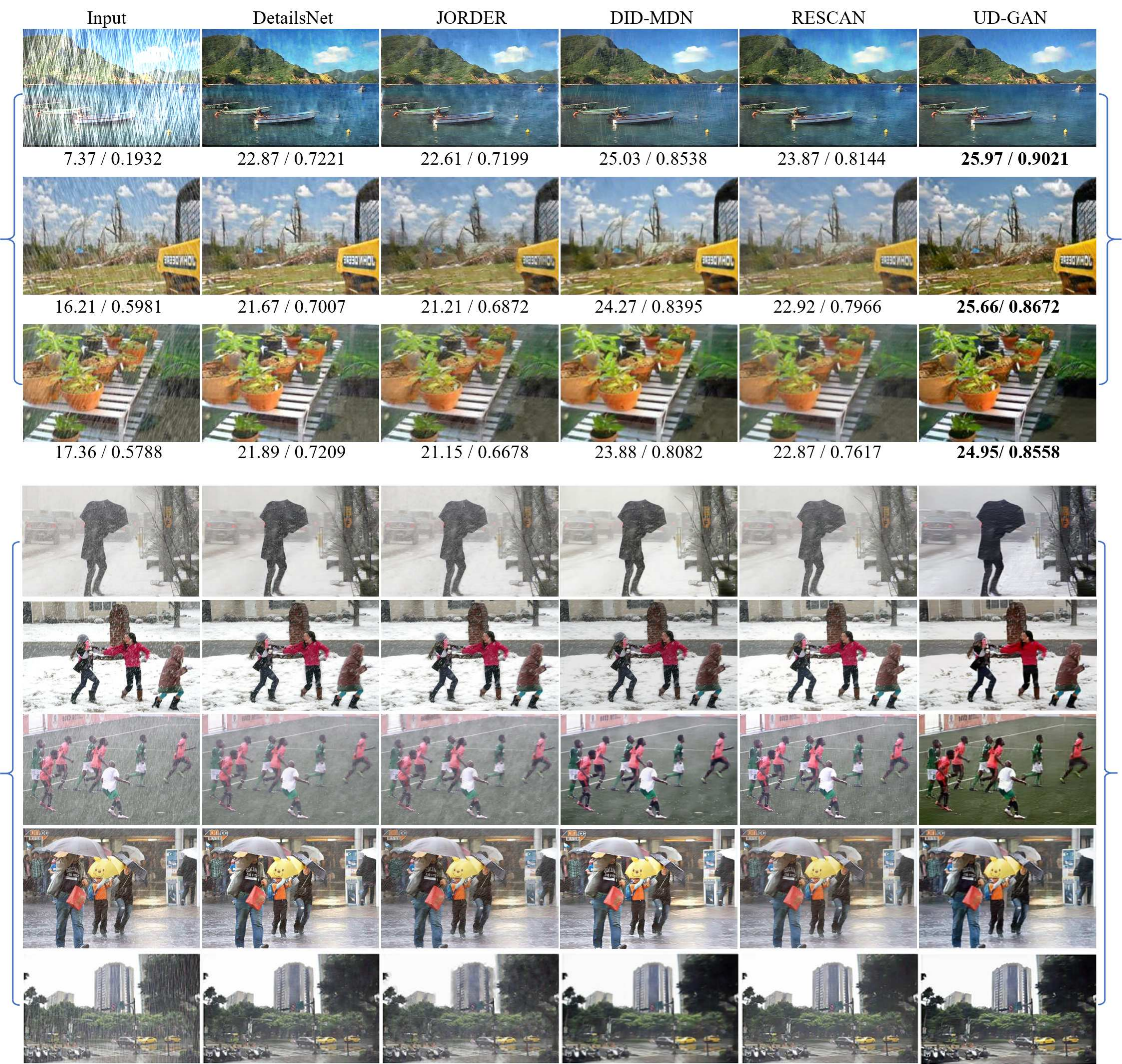}}
  \caption{Deraining results of different methods on rainy images from the synthetic datasets (top three rows), and real-world dataset (bottom five rows). PSNR/SSIM have been calculated and attached below synthetic samples.}
  \centering
\label{fig:Syn-Real-results}
\end{figure*}

\begin{figure*}
  \centerline{\includegraphics[width=1.0\linewidth]{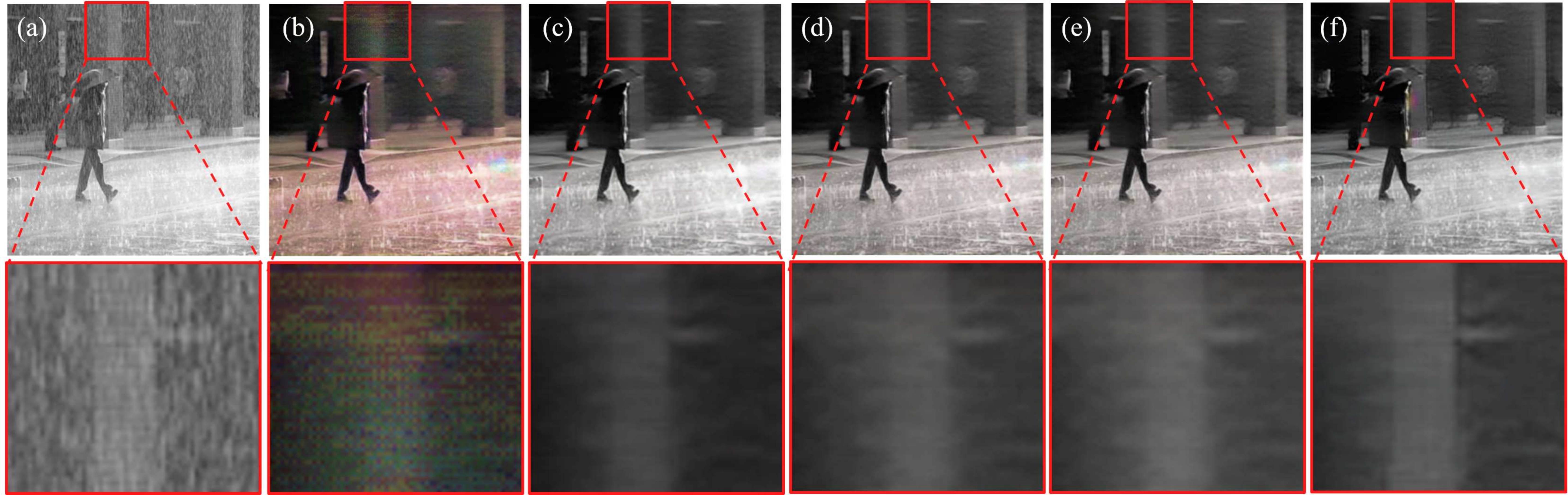}}
  \caption{Subjective results of removing different components of UD-GAN. From left to right: (a) rainy input, (b) Ours - \{RGM,BGM,lum\}, (c) Ours - RGM, (d) Ours - BGM, (e) Ours - lum, and (f) the complete UD-GAN.}
  \centering
\label{fig:AblationStudy}
\end{figure*}

\subsection{Comparison Results}

\textbf{Synthetic Data:} Table \ref{Quantitative Evaluation} shows the quantitative results of different methods on synthetic datasets Rain800, Rain12, Rain100L and Rain100H. We can observe that our method is able to perform equally well when only training on the synthetic datasets (UD-GAN$_{syn}$) or real-world datasets (UD-GAN$_{real}$) compared to other fully-supervised methods, and even outperform them (over 0.7dB PSNR gain) when training on both synthetic and real-world datasets. That is because (1) the involvement of real-world rainy images helps our model break away from the limitation of synthetic data, reaching the effect of data augmentation, (2) based on the reliable intrinsic statistics of unpaired rainy and clean images, we provide more generalized, appropriate and reasonable self-supervised constraints for the network than existing fully-supervised methods.

It can be noted that, although the PSNR and SSIM of UD-GAN$_{real}$ on some synthetic datasets is lower than that of the best STOA supervised solution DID-MDN \cite{zhang2018density}, but the subjective de-rained effect of UD-GAN$_{real}$ is obviously better than DID-MDN as shown in Figure \ref{fig:real}.

In Figure \ref{fig:Syn-Real-results} (top three rows), we select most advanced methods and most difficult synthetic rainy images to further show that UD-GAN promises the most satisfactory subjective de-rained effect, which effectively removes rain steaks while preserving better details.

\textbf{Real-world Data:} To test the practicability of UD-GAN, we also evaluate its performance on real-world rainy images. Figure \ref{fig:Syn-Real-results} (bottom five rows) shows some de-rained results on the real-world test dataset: DetailsNet \cite{fu2017removing} tends to leave residual rain streaks in the background. DID-MDN \cite{zhang2018density} over-smooths some important details such as building structures as shown in the last row, and cannot handle these snow-like raindrops as shown in the first, second rows. JORDER \cite{yang2017deep} and RESCAN \cite{li2018recurrent} suffer from unexpected artifacts on the de-rained results as shown in the middle and last row (please zooming-in to observe). In contrast, UD-GAN effectively restores clean background with rich texture details while promising more natural and realistic luminance, which significantly improves the subjective effects and greatly surpasses other methods in terms of clarity and visibility.

\begin{table}
\scriptsize
\centering
\caption{Objective results of removing different components of UD-GAN.}
\scalebox{1}{
\label{Ablation Study_1}
\begin{tabular}{|c|c|c|c|c|c|}
\hline
         & \multicolumn{5}{c|}{PSNR}                                                                                                                                                                                                                                                                                             \\ \hline
Methods  & \begin{tabular}[c]{@{}c@{}}Ours \\ - \{RGM,BGM,lum\}\end{tabular} & \begin{tabular}[c]{@{}c@{}}Ours \\ - RGM\end{tabular} & \begin{tabular}[c]{@{}c@{}}Ours \\ - BGM\end{tabular} & \begin{tabular}[c]{@{}c@{}}Ours\\ - lum\end{tabular} & \textbf{\begin{tabular}[c]{@{}c@{}}Ours\end{tabular}} \\ \hline
Rain800  & 23.58                                                & 24.85                                                  & 25.07                                                                           & 25.77                                                  & \textbf{25.98}                                             \\ \hline
Rain12   & 34.61                                                & 35.13                                                  & 35.31                                                                           & 36.81                                                  & \textbf{37.13}                                             \\ \hline
Rain100L & 34.67                                                & 35.89                                                  & 35.86                                                                           & 36.72                                                  & \textbf{37.28}                                             \\ \hline
Rain100H & 23.72                                                & 25.36                                                  & 25.62                                                                           & 27.14                                                  & \textbf{27.75}                                             \\ \hline
         & \multicolumn{5}{c|}{SSIM}                                                                                                                                                                                                                                                                                             \\ \hline
Rain800  & 0.8133                                               & 0.8749                                                 & 0.8792                                                                          & 0.8926                                                 & \textbf{0.9093}                                            \\ \hline
Rain12   & 0.9421                                               & 0.9553                                                 & 0.9501                                                                          & 0.9591                                                 & \textbf{0.9644}                                            \\ \hline
Rain100L & 0.9379                                               & 0.9525                                                 & 0.9626                                                                          & 0.9692                                                 & \textbf{0.9753}                                            \\ \hline
Rain100H & 0.7249                                               & 0.8359                                                 & 0.8323                                                                          & 0.8697                                                 & \textbf{0.8931}                                            \\ \hline
\end{tabular}}
\end{table}

\subsection{Ablation Study}

To verify the effectiveness of the proposed Rain Guidance Module (RGM), Background Guidance Module (BGM) and luminance-adjusting adversarial loss in UD-GAN, we implement four ablated schemes for comparison. Due to space limitation, here we abbreviate the complete UD-GAN as \textbf{``Ours''}:

\begin{itemize}
    \setlength{\itemsep}{0pt}
    \setlength{\parsep}{0pt}
    \setlength{\parskip}{0pt}
    \item {\textbf{Ours - \{RGM,BGM,lum\}}}. Benchmark scheme without RGM, BGM or luminance-adjusting loss.
    \item {\textbf{Ours - RGM}}. Remove Rain Guidance Module.
    \item {\textbf{Ours - BGM}}. Remove Background Guidance Module.
    \item {\textbf{Ours - lum}}. Remove luminance-adjusting component in the adversarial loss.
\end{itemize}

Table \ref{Ablation Study_1} shows that the complete UD-GAN achieves 2.40/2.52/2.61/4.03dB, 1.13/2.00/1.39/2.39dB, 0.91/1.82/1.42/2.13dB and 0.21/0.32/0.56/0.61dB PSNR gain over four ablated baselines (Ours - \{RGM,BGM,lum\}, Ours - RGM, Ours - BGM and Ours - lum) on Rain800, Rain12, Rain100L and Rain100H respectively.

% Although the luminance-adjusting loss doesn't bring a large gain on PSNR, the improvement of subjective effect is obvious.

Subjective comparisons are presented in Figure \ref{fig:AblationStudy}: Benchmark scheme produces unacceptable artifacts. RGM makes the restored background more clear and visible. BGM promises more realistic and relatively richer textural content. Luminance-adjusting adversarial loss helps the de-rained output has natural luminance and looks cleaner.

\subsection{User Study}

For a more comprehensive qualitative evaluation, we conduct three user studies to respectively demonstrate the effectiveness of UD-GAN$_{syn}$, UD-GAN$_{real}$ and UD-GAN in generating visually attractive results. To build the subjective database, we choose 12 rainy images randomly from synthetic and real-world test data with a ratio of 1:1, and each image is de-rained by DetailsNet \cite{fu2017removing}, JORDER \cite{yang2017deep}, DID-MDN \cite{zhang2018density}, RESCAN \cite{li2018recurrent}, UD-GAN$_{syn}$, UD-GAN$_{real}$ and UD-GAN separately. Then we respectively compare the de-rained results of UD-GAN$_{syn}$, UD-GAN$_{real}$ and UD-GAN with other 4 methods in three independent user studies (No.1, No.2, No.3). Refer to the subjective experiment design \cite{li2018recurrent} and assessment criteria \cite{bt2002methodology,ponomarenko2015image} in the previous studies, we show the original rainy image together with its five de-rained results using different methods to 10 non-expert subjects, then they are instructed to vote for the best de-rained result with the least rain streaks and the clearest texture details. As shown in Table \ref{Subjective Experiment}, UD-GAN$_{syn}$, UD-GAN$_{real}$ and UD-GAN respectively gets the most 88, 93, 101 votes in three independent user studies, which demonstrates the superiority of our method in synthesizing subjective high-quality de-rained images.

\subsection{Time Complexity Comparisons}

Computational time comparisons are shown in Table \ref{Time}. The proposed UD-GAN is comparable to other methods because only $G_c$ works when testing, it only takes about 0.25s on average to process a rainy image of size 512*512.

\subsection{Extension}

We validate that UD-GAN can generalize to other low-level image processing tasks such as \textbf{image denoising}, and also can be used to pre-process for high-level vision such as \textbf{action recognition}. Experimental results can be found in Table \ref{Denoising}.

\begin{table}
\centering
\scriptsize
\caption{Results of three user studies (Voting number of different methods, the higher the better).}
\scalebox{0.9}{
\label{Subjective Experiment}
\begin{tabular}{|c|c|c|c|c|c|c|c|}
\hline
& DetailsNet & JORDER & \begin{tabular}[c]{@{}c@{}}DID-\\ MDN\end{tabular} & RESCAN & \textbf{\begin{tabular}[c]{@{}c@{}}UD-\\ GAN\\$_{syn}$\end{tabular}} & \textbf{\begin{tabular}[c]{@{}c@{}}UD-\\ GAN\\$_{real}$\end{tabular}} & \textbf{\begin{tabular}[c]{@{}c@{}}UD-\\ GAN\end{tabular}} \\ \hline
No.1 &0          & 2      & 24      & 6      & 88      & --      & --     \\ \hline
No.1 &2          & 1      & 23      & 1      & --      & 93      & --     \\ \hline
No.3 &1          & 1      & 14      & 3      & --      & --      & 101    \\ \hline
\end{tabular}}
\end{table}

\begin{table}
\centering
\scriptsize
\caption{Computational time for different methods averaged on 200 images with size 512*512.}
\label{Time}
\scalebox{0.93}{
\begin{tabular}{|c|c|c|c|c|c|}
\hline
                                                        & DetailsNet & JORDER & \begin{tabular}[c]{@{}c@{}}DID-\\ MDN\end{tabular} & RESCAN & \textbf{\begin{tabular}[c]{@{}c@{}}UD-\\ GAN\end{tabular}} \\ \hline
\begin{tabular}[c]{@{}c@{}}512*512\\ (GPU)\end{tabular} & 0.34s      & 1.88s  & 0.28s                                             & 4.75s  & \textbf{0.25s}                                             \\ \hline
\end{tabular}}
\end{table}

% \usepackage{booktabs}
\begin{table}
\centering
\scriptsize
\caption{Denoising results (SSIM) on BSD68 \cite{roth2009fields}.}
\label{Denoising}
\begin{tabular}{|c|c|c|c|} \hline
BSD68 \cite{roth2009fields}  & $\sigma$=15    & $\sigma$=25    & $\sigma$=50    \\ \hline
\hline
DnCNN \cite{zhang2017beyond}  & 0.8826 & 0.8190 & 0.7076 \\
UD-GAN & 0.8822 & 0.8198 & 0.7088 \\ \hline
\end{tabular}
\end{table}

\section{Conclusion}

In this paper, we tackle the single image deraining problem in an unsupervised manner with an end-to-end learned model, i.e. Unsupervised Deraining GAN (UD-GAN). Compared to existing supervised approaches which attempt to learn a mapping between synthetic rainy inputs and corresponding ground truths, we provide the network with more reliable and reasonable self-supervised constraints from the intrinsic statistics of original data through two collaboratively optimized modules BGM \& RGM and a luminance-adjusting adversarial loss. Sufficient experiments and comparisons are performed on both synthetic and real-world datasets to demonstrate the effectiveness, generalization and practicability of the proposed UD-GAN. In future, we plan to extend the proposed self-supervision algorithm to a wider range of unsupervised image restoration tasks including deblurring, dehazing and super-resolution.

{\small
\bibliographystyle{ieee}
\bibliography{egbib}
}

\end{document}